\documentclass[10pt,conference]{IEEEtran}

\usepackage{booktabs}
\usepackage{multirow}
\usepackage{pgfplots}
\pgfplotsset{compat=1.18}
\usepgfplotslibrary{groupplots}
\usetikzlibrary{positioning}
\usepackage[numbers,sort&compress]{natbib}
\usepackage{url}
\usepackage{amsmath}

\title{98$\times$ Faster LLM Routing Without a Dedicated GPU: Flash Attention, Prompt Compression, and Near-Streaming for the vLLM Semantic Router}

\author{
  Xunzhuo Liu$^{1}$, Bowei He$^{2,3, \dagger}$, Xue Liu$^{2,3}$, Andy Luo$^{4}$, Haichen Zhang$^{4}$, Huamin Chen$^{5}$ \\
  $^1$ vLLM Semantic Router Project, $^2$ MBZUAI, $^3$ McGill University, $^4$ AMD, $^5$ Red Hat \\
  $\dagger$ Corresponding author: \texttt{Bowei.He@mbzuai.ac.ae}
}

\begin{document}

\maketitle

\begin{abstract}
System-level routers that intercept LLM requests for safety classification, domain routing, and PII detection must be both fast and operationally lightweight: they should add minimal latency to every request, yet not require a dedicated GPU---an expensive resource better used for LLM inference itself. When the router co-locates on the same GPU as vLLM serving instances, standard attention's $O(n^2)$ memory makes long-context classification (8K--32K tokens) impossible: at 8K tokens, three concurrent classifiers need ${\sim}$4.5\,GB for attention masks alone, far exceeding the memory left by vLLM. We present three staged optimizations for the vLLM Semantic Router, benchmarked on AMD Instinct MI300X, that solve both the latency and the memory problem. \emph{Stage~1}: a custom CK Flash Attention operator for ONNX Runtime on ROCm reduces attention memory from $O(n^2)$ to $O(n)$ and end-to-end (E2E) latency from 4{,}918\,ms to 127\,ms (\textbf{38.7$\times$}), enabling 8K--32K tokens where SDPA OOMs. \emph{Stage~2}: classical NLP prompt compression (TextRank, position weighting, TF-IDF, and novelty scoring) reduces all inputs to ${\sim}$512 tokens without neural inference, capping both latency and GPU memory at a constant regardless of original prompt length (E2E 127$\to$62\,ms, \textbf{2.0$\times$}). \emph{Stage~3}: near-streaming body processing with adaptive chunking and zero-copy JSON eliminates serialization overhead (E2E 62$\to$50\,ms, \textbf{1.2$\times$}). Cumulatively: \textbf{98$\times$} improvement (4{,}918\,ms to 50\,ms), 16K-token routing in 108\,ms, and a total router GPU footprint under 800\,MB---small enough to share a GPU with LLM serving and removing the need for a dedicated accelerator. Stage~1 targets AMD ROCm (NVIDIA GPUs already have FlashAttention via cuDNN); Stages~2 and~3 are hardware-agnostic. \footnote{Model, benchmark, and code are provided in: \url{https://github.com/vllm-project/semantic-router} and \url{https://huggingface.co/llm-semantic-router}.}
\end{abstract}

\section{Introduction}
\label{sec:intro}

Large Language Model (LLM) inference serving systems such as vLLM~\citep{kwon2023vllm} increasingly rely on system-level routing layers to enforce safety policies, classify intent, detect personally identifiable information (PII), and select optimal backend models before requests reach GPU inference endpoints. The vLLM Semantic Router~\citep{vllmsr2025} implements this as an Envoy~\citep{envoy2024} external processing (ext\_proc) filter: a gRPC service that intercepts HTTP requests in the proxy data plane, evaluates multiple classification signals in parallel, and mutates headers to route requests to appropriate endpoints.

This architecture imposes a tight latency budget. Every millisecond spent in the routing layer adds directly to the user-perceived time-to-first-token (TTFT). For short prompts ($\leq$512 tokens), GPU-based classification with ModernBERT~\citep{modernbert2024} completes in 9--14\,ms per classifier (Table~\ref{tab:gpu-results}). However, real-world prompts are growing: retrieval-augmented generation (RAG) pipelines, code review, and document summarization routinely produce 8K--32K token contexts. At these lengths, three bottlenecks emerge:

\begin{enumerate}
\item \textbf{Quadratic attention complexity.} Standard scaled dot-product attention (SDPA) materializes an $O(n^2)$ attention mask. At 8K tokens a single classifier session allocates ${\sim}$1.5\,GB for this mask; at 16K the mask grows to ${\sim}$6\,GB; at 32K, ${\sim}$24\,GB. Because the router co-locates on the same GPU as vLLM inference instances---which consume the majority of HBM for KV cache and model weights---only ${\sim}$718\,MB remains per classifier session. With three concurrent classifiers multiplying the demand, SDPA OOMs at 8K+ tokens. Higher batch concurrency (multiple simultaneous requests) further multiplies memory pressure, making the problem worse under production load.
\item \textbf{CPU inference scaling.} CPU-based ONNX and Candle backends avoid GPU memory pressure but show superlinear latency growth: 8K tokens require 4.9\,s (ONNX) or 1.8\,s (Candle, truncated to 512) for a single request through the full pipeline---far exceeding any acceptable routing budget.
\item \textbf{Proxy serialization overhead.} Envoy's \texttt{BUFFERED} body mode requires the entire HTTP body to be accumulated, then serialized as a single protobuf message. The ext\_proc service must deserialize it, perform \texttt{json.Unmarshal} on the full OpenAI-format payload, and \texttt{json.Marshal} the mutated body---operations that scale linearly with payload size.
\end{enumerate}

We address each bottleneck with a targeted optimization and show that the three stages compose multiplicatively at 8K tokens: GPU+FA alone delivers 38.7$\times$ (4{,}918$\to$127\,ms), adding prompt compression yields 79$\times$ (127$\to$62\,ms), and adding near-streaming reaches \textbf{98$\times$} (62$\to$50\,ms). At 16K tokens the combined system achieves 108\,ms end-to-end---a regime where CPU backends cannot operate. While Stage~1 targets AMD ROCm (NVIDIA GPUs already have FlashAttention via cuDNN and libraries such as Candle), Stages~2 and~3 are hardware-agnostic and benefit any GPU or CPU backend. Our contributions are:

\begin{enumerate}
\item \textbf{CK Flash Attention for ONNX Runtime on ROCm.} A custom operator, ONNX graph rewriter, and HIP kernel that bring FlashAttention to ONNX Runtime on AMD GPUs for the first time, reducing classifier memory from $O(n^2)$ to $O(n)$ allowing 8K--32K token sequences where SDPA OOMs (\S\ref{sec:gpu}).
\item \textbf{Neural-inference-free prompt compression.} A classical NLP pipeline combining TextRank, U-shaped position weighting, TF-IDF, and novelty scoring that compresses prompts to ${\sim}$512 tokens before classification with no model call and no accuracy loss, applicable to any backend (\S\ref{sec:compression}).
\item \textbf{Near-streaming body processing.} An adaptive chunked handler for Envoy ext\_proc that detects the routing mode from the first chunk and selects between zero-copy passthrough and incremental accumulation, avoiding full-body JSON serialization for the majority of requests (\S\ref{sec:streaming}).
\item \textbf{GPU co-location with LLM inference.} The combined memory and latency reductions let the router share a GPU with vLLM serving instances, removing the need for a dedicated accelerator and improving GPU density in production clusters (\S\ref{sec:discussion}).
\end{enumerate}

\section{Background and Related Work}
\label{sec:background}

\subsection{Envoy External Processing}

\begin{sloppypar}
Envoy's ext\_proc filter~\citep{envoy2024} enables out-of-process request interception via bidirectional gRPC streaming. The router receives \texttt{ProcessingRequest} messages for headers, body, and trailers, and responds with mutations (header modifications, body rewrites, or immediate responses). In \texttt{BUFFERED} mode, the entire request body is delivered in a single message; in \texttt{STREAMED} mode, the body arrives as a sequence of fixed-size chunks with an end-of-stream flag on the final chunk.
\end{sloppypar}

\subsection{Attention Mechanisms and Flash Attention}

\begin{sloppypar}
Standard self-attention computes $\mathrm{softmax}(QK^T\!/\!\sqrt{d_k})\,V$, requiring $O(n^2)$ memory for the attention matrix. FlashAttention~\citep{dao2022flashattention} restructures this via tiling and kernel fusion to achieve $O(n)$ memory with no approximation error. FlashAttention-2~\citep{dao2023flashattention2} further improves parallelism. AMD's Composable Kernel (CK) library~\citep{amdck2024} provides a ROCm-native implementation using tiled fused multi-head attention (FMHA) kernels optimized for MI300X.
\end{sloppypar}

\subsection{Prompt Compression}

\citet{liu2024lost} demonstrated that LLMs exhibit a U-shaped attention pattern: information at the beginning and end of the context receives disproportionate attention, while the middle is ``lost.'' This observation informs position-aware prompt compression strategies.

Recent prompt compression methods fall into three categories. \emph{LLM-based} methods such as LLMLingua~\citep{jiang2023llmlingua} use a small causal language model (e.g., LLaMa-7B) to compute per-token perplexity and prune low-entropy tokens, reaching up to 20$\times$ compression with minimal task performance loss. LLMLingua-2~\citep{pan2024llmlingua2} improves on this by framing compression as token classification using a bidirectional encoder (XLM-RoBERTa) trained on a distilled compression dataset, running 3--6$\times$ faster than LLMLingua. \emph{Self-information} methods like Selective Context~\citep{li2023selective} use a base LM to score lexical units by self-information, retaining only high-information segments. \emph{Learned abstractive} methods such as RECOMP~\citep{xu2024recomp} train compressors end-to-end for RAG settings, generating summaries at 6\% compression rate.

All of these approaches require LLM or neural network inference at compression time---precisely the overhead we need to avoid in a latency-sensitive routing layer. TextRank~\citep{mihalcea2004textrank}, a graph-based unsupervised ranking algorithm inspired by PageRank, provides an extractive method with no neural inference. Combined with TF-IDF weighting~\citep{sparckjones1972idf}, position-aware scoring, and centroid-based novelty detection adapted from \citet{radev2004centroid} and the diversity principle of MMR~\citep{carbonell1998mmr}, these classical NLP techniques can compress prompts without any model call. Table~\ref{tab:compression-comparison} compares the approaches.

\begin{table*}[t]
\caption{Comparison of prompt compression approaches. Compression latency is for a 16K-token input. Our method is the only one requiring zero neural inference, keeping compression overhead to ${\sim}$19\,ms at 16K tokens vs.\ seconds for neural methods.$^{\dagger}$}
\label{tab:compression-comparison}
\centering
\footnotesize
\begin{tabular}{lcccc}
\toprule
\textbf{Method} & \textbf{Requires Neural Inf.} & \textbf{Compression Granularity} & \textbf{Attention Preserved?} & \textbf{Compression Latency} \\
\midrule
LLMLingua~\citep{jiang2023llmlingua} & LLaMa-7B & Token & Yes$^*$ & ${\sim}$2--5\,s \\
LLMLingua-2~\citep{pan2024llmlingua2} & XLM-R (560M) & Token & Yes$^*$ & ${\sim}$0.3--1\,s \\
Selective Context~\citep{li2023selective} & GPT-2/LLaMa & Sentence/phrase & Partial & ${\sim}$1--3\,s \\
RECOMP~\citep{xu2024recomp} & Trained encoder & Abstractive & No (rewritten) & ${\sim}$0.5--2\,s \\
\midrule
\textbf{Ours} & \textbf{None} & Sentence & \textbf{Yes (U-shaped + novelty)} & \textbf{${\sim}$19\,ms} \\
\bottomrule
\end{tabular}
\\[2pt]
{\scriptsize $^*$Attention preservation via perplexity scoring, not explicit position weighting.
$^{\dagger}$Latencies for other methods are estimates based on model sizes and reported compression rates; our 19\,ms is measured directly on the compression pipeline at 16K tokens (Table~\ref{tab:compression-ratio}).}
\end{table*}

\subsection{ModernBERT}

ModernBERT~\citep{modernbert2024} is a bidirectional encoder trained on 2 trillion tokens with native 8K context length, rotary position embeddings (RoPE), and alternating local/global attention layers. ModernBERT natively supports FlashAttention: its alternating local (128-token sliding window) and global attention pattern maps directly to FA's tiled kernel interface. We use a fine-tuned variant (mmBERT-32K) with YaRN RoPE scaling~\citep{peng2023yarn} extended to 32K tokens, deployed as three concurrent ONNX sessions for domain classification, jailbreak detection, and PII detection.

\section{System Architecture}
\label{sec:architecture}

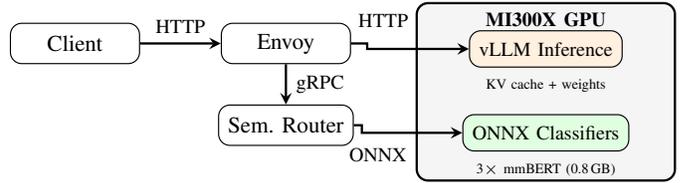
\begin{figure}[t]
\centering
\resizebox{\columnwidth}{!}{%
\begin{tikzpicture}[
  box/.style={rectangle, draw, rounded corners, minimum width=1.6cm, minimum height=0.5cm, font=\footnotesize},
  gpubox/.style={rectangle, draw, rounded corners, fill=#1, minimum width=1.4cm, minimum height=0.45cm, font=\scriptsize},
  gpubox/.default={blue!8},
  arrow/.style={->, thick, >=stealth},
]
\node[box] (client) {Client};
\node[box, right=1.0cm of client] (envoy) {Envoy};
\node[box, below=0.5cm of envoy] (router) {Sem.\ Router};

\node[draw, thick, rounded corners, fill=gray!8,
      minimum width=3.2cm, minimum height=2.2cm,
      right=0.8cm of envoy, yshift=-0.6cm,
      label={[font=\scriptsize\bfseries, anchor=north]north:MI300X GPU}] (gpu) {};

\node[gpubox=orange!12, anchor=north] at ([yshift=-0.35cm]gpu.north) (vllm) {vLLM Inference};
\node[font=\tiny, below=0.02cm of vllm] (vllmmem) {KV cache + weights};
\node[gpubox=green!12, below=0.15cm of vllmmem] (onnx) {ONNX Classifiers};
\node[font=\tiny, below=0.02cm of onnx] (onnxmem) {3$\times$ mmBERT (0.8\,GB)};

\draw[arrow] (client) -- node[above, font=\scriptsize] {HTTP} (envoy);
\draw[arrow] (envoy) -- node[right, font=\scriptsize] {gRPC} (router);
\draw[arrow] (envoy.east) -- ++(0.2,0) |- node[above, near start, font=\scriptsize, xshift=0.2cm, yshift=0.1cm] {HTTP} (vllm.west);
\draw[arrow] (router.east) -- ++(0.1,0) |- node[below, near start, font=\scriptsize, xshift=0.2cm, yshift=-0.1cm] {ONNX} (onnx.west);
\end{tikzpicture}%
}
\caption{Co-located deployment: the router's ONNX classifiers share the MI300X GPU with vLLM inference, consuming only ${\sim}$0.8\,GB vs.\ vLLM's ${\sim}$190\,GB---no dedicated accelerator needed.}
\label{fig:architecture}
\end{figure}

The vLLM Semantic Router operates as an Envoy ext\_proc filter in the data plane of a Kubernetes-based LLM serving infrastructure. The router's classifier sessions share the same GPU as one or more vLLM inference instances rather than requiring a dedicated GPU. Dedicating an entire MI300X (192\,GB HBM3) solely to lightweight classification would be wasteful. However, it means the router must operate within whatever GPU memory the vLLM instances leave available---typically a small fraction of total HBM. Figure~\ref{fig:architecture} illustrates the request flow.

The router processes each request through a pipeline of stages: (1)~header inspection and model detection, (2)~body parsing and content extraction, (3)~prompt compression (optional), (4)~parallel signal evaluation (jailbreak, PII, domain, modality), (5)~decision engine evaluation, and (6)~model routing and header mutation.

\section{System Design}
\label{sec:design}

The vLLM Semantic Router addresses each bottleneck from \S\ref{sec:intro} with a dedicated optimization layer. This section describes the design of all three stages; quantitative evaluation follows in \S\ref{sec:evaluation}.

\subsection{CK Flash Attention for ONNX Runtime on ROCm}
\label{sec:gpu}

\subsubsection{ONNX Runtime with ROCm Execution Provider}

We compile the mmBERT-32K classifiers to ONNX format and execute them through ONNX Runtime~\citep{onnxruntime2021} with the ROCm execution provider on AMD Instinct MI300X GPUs (192GB HBM3, gfx942 architecture). Three classifier sessions (domain, jailbreak, PII) run concurrently, sharing the GPU memory pool.

\subsubsection{Why Flash Attention Is Unavailable in ONNX Runtime for ROCm}

ONNX Runtime includes built-in FlashAttention support only for NVIDIA CUDA execution providers, specifically targeting SM~80--89 architectures via operators such as \texttt{com.microsoft.\allowbreak Multi\-Head\-Attention} and \texttt{Group\-Query\-Attention}. These kernels call cuDNN or the Dao-AILab flash-attention library~\citep{dao2022flashattention}, neither of which runs on AMD GPUs. The ROCm execution provider in ORT falls back to standard SDPA, which materializes the full $[1,1,S,S]$ attention mask in HBM. At 8K tokens, a single session requests a 1.5\,GB buffer for this mask. Because the router shares the GPU with vLLM inference instances (which consume the majority of HBM for KV cache and model weights), only ${\sim}$718\,MB remains available per classifier session, and allocation fails.

Although AMD provides Flash Attention through the Composable Kernel (CK) library~\citep{amdck2024} for PyTorch (merged as a CK backend in late 2024), no equivalent integration exists for ONNX Runtime. The ORT custom operator API is the only extensibility path, and it requires solving three additional problems: (1)~defining the ONNX graph transformation to replace SDPA sub\-graphs, (2)~mapping ModernBERT's alternating local/global attention pattern to CK's sliding-window parameters, and (3)~converting the $O(n^2)$ 2-D attention mask to an $O(n)$ 1-D padding bias compatible with CK's kernel interface.

\subsubsection{Custom CK Flash Attention Operator}

We implement a custom ORT operator registered under the \texttt{com.ck} domain that bridges this gap. The system has three components:

\paragraph{ONNX Graph Rewrite.}
A Python script (\texttt{rewrite\_graph.py}) performs pattern-matching on the exported ONNX graph to locate SDPA subgraphs. The pattern detected is:
\begin{quote}\small
\texttt{Q*scale $\to$ MatMul(Q,K\textsuperscript{T}) $\to$ Add(mask)}\\\texttt{$\to$ Softmax $\to$ MatMul(attn,V)}
\end{quote}
\begin{sloppypar}
Each matched subgraph is replaced by a single \texttt{CKFlash\-Attention} node with inputs \texttt{[Q, K, V, pad\_bias]} and attributes \texttt{scale}, \texttt{window\_size\_left}, and \texttt{window\_size\_right}. The 2-D attention mask $[B,H,S_q,S_k]$ (from \texttt{Where}/\texttt{Expand} nodes) is replaced by a 1-D padding bias $[B,1,1,S]$ derived from the input \texttt{attention\_mask}:
\end{sloppypar}
\begin{equation}
\texttt{pad\_bias} = -65504 \cdot (1 - \texttt{attention\_mask})
\end{equation}
cast to FP16. A dead-code elimination pass then removes the now-unused 2-D mask construction subgraph, recovering the $O(n^2) \to O(n)$ memory savings.

\paragraph{ModernBERT Attention Mapping.}
ModernBERT uses alternating local and global attention layers: local layers apply a sliding window of 128 tokens, while global layers (every 3rd layer) attend to the full sequence. The rewrite script detects local vs.\ global layers by mask node names (e.g., \texttt{Where\_1} vs.\ \texttt{Where\_2}) and assigns per-layer window parameters:
\begin{itemize}
\item Local layers: \texttt{window\_left=63, window\_right=64} (128-token window)
\item Global layers: \texttt{window\_left=$-$1, window\_right=$-$1} (full attention)
\end{itemize}

\paragraph{HIP Kernel and ORT Registration.}
\begin{sloppypar}
The C++/HIP implementation (\texttt{ort\_custom\_op.cpp} + \texttt{ck\_fmha\_dispatch.hip}) registers \texttt{CKFlashAttention} as an ORT custom operator bound to the \texttt{ROCMExecutionProvider}. At compute time, the kernel reads Q, K, V tensors and the optional padding bias, obtains the HIP stream from ORT's kernel context, and dispatches to CK-tile's \texttt{FmhaFwdFp16} kernel with \texttt{GenericAttentionMask} for sliding-window support. Runtime head-dimension dispatch covers 32, 64, and 128. The compiled shared library (\texttt{libort\_ck\_flash\_attn.so}, 192\,KB, targeting \texttt{gfx942}) is loaded at session creation via \texttt{ORT\_CK\_FLASH\_ATTN\_LIB}.
\end{sloppypar}

\paragraph{Model Preparation Pipeline.}
The end-to-end flow is:
\begin{enumerate}\begin{sloppypar}
\item Export mmBERT-32K from PyTorch to ONNX (\texttt{model\_sdpa\_fp16.onnx}).
\item Run \texttt{rewrite\_graph.py} with \texttt{-{}-hdim~64} \texttt{-{}-local-attention~128} to produce the FA model.
\item At runtime, the Rust ONNX binding checks for \texttt{ORT\_CK\_FLASH\_ATTN\_LIB} and loads the FA variant.
\end{sloppypar}\end{enumerate}
This pipeline is fully automated in the CI/CD Dockerfile, and prebuilt FA models are published to HuggingFace for deployment.

\subsection{Prompt Compression for Classification}
\label{sec:compression}

Flash Attention solves the OOM problem by reducing attention memory from $O(n^2)$ to $O(n)$, enabling 8K--32K token classification on a shared GPU. However, two scaling challenges remain. First, FA \emph{latency} still grows with sequence length: 259\,ms per classifier at 16K tokens, 756\,ms at 32K (Table~\ref{tab:sdpa-fa}). Under concurrent load, FA memory also accumulates---at C=20 with 32K tokens, the combined working set becomes substantial even at $O(n)$ per request. Second, GPU contention from three concurrent classifier sessions further increases wall-clock time. We observe that classification signals (jailbreak intent, PII presence, domain category) are typically concentrated in specific regions of the prompt, not uniformly distributed. We therefore compress the prompt \emph{before} classification: by reducing all inputs to ${\sim}$512 tokens regardless of original length, compression simultaneously caps FA latency at ${\sim}$19\,ms, caps per-request GPU memory at a constant small footprint, and decouples both metrics from the original prompt length.

\subsubsection{Algorithm}

Our compression pipeline scores each sentence with four signals, then selects the top-ranked sentences to fill the token budget. Following MEAD~\citep{radev2004centroid}, which pioneered the weighted-combination approach to extractive summarization (centroid $+$ position $+$ first-sentence overlap), we combine four complementary signals in a linear composite. All pairwise similarities use the cosine measure on term-frequency vectors~\citep{salton1988term}.

\paragraph{Sentence Segmentation.}
The input text is split into sentences using a rule-based multilingual segmenter supporting Latin scripts, CJK characters, Arabic, and Devanagari. For GC efficiency with large prompts, sentences are capped at 500 before ranking.

\paragraph{Signal 1: TextRank (Centrality).}
Following \citet{mihalcea2004textrank}, we construct a cosine-similarity graph over sentence TF vectors and compute PageRank scores via power iteration. Sentences similar to many other important sentences score high. We use a flat adjacency matrix with \texttt{sync.Pool} recycling to avoid GC pressure on large inputs.

\paragraph{Signal 2: Position Weighting.}
\citet{liu2024lost} demonstrated that LLM attention follows a U-shaped curve over input position. We parametrize this empirical finding as:
\begin{equation}
w_{\mathrm{pos}}(i) = 1 - d \cdot \sin\!\left(\frac{\pi \, i}{n-1}\right)
\end{equation}
where $i$ is the 0-indexed sentence position, $n$ is the total sentence count, and $d \in [0,1]$ controls the curve depth (default $d{=}0.5$). The sine function is the natural single-parameter smooth approximation of the reported U-shape: at $d{=}0.5$, edge sentences receive weight~1.0 and the middle sentence 0.5.

\paragraph{Signal 3: TF-IDF Information Density.}
Each sentence receives a mean TF-IDF score~\citep{sparckjones1972idf} computed over the document's sentence corpus. Sentences containing rare, discriminative terms---those with high inverse document frequency---score higher, following the self-information approximation of Selective Context~\citep{li2023selective} but without neural inference.

\paragraph{Signal 4: Novelty (Inverse Centrality).}
TextRank rewards centrality, but safety-critical content (jailbreak prefixes, PII) is typically an \emph{outlier}. Adapting centroid-based scoring~\citep{radev2004centroid} in the opposite direction, we score each sentence by its dissimilarity from the document centroid:
\begin{equation}
r_{\mathrm{nov}}(i) = 1 - \cos(\mathbf{tf}_i,\, \bar{\mathbf{tf}})
\end{equation}
where $\mathbf{tf}_i$ is the sentence's TF vector and $\bar{\mathbf{tf}}$ is the mean of all sentence TF vectors. Where \citet{radev2004centroid} retain sentences closest to the centroid (most representative), we boost sentences farthest from it, following the diversity principle of MMR~\citep{carbonell1998mmr}. Adversarial or sensitive content scores high without keyword lists or pattern matching---purely from distributional divergence.

\paragraph{Composite Ranking and Selection.}
The final score for sentence $i$ is:
\begin{equation}
\begin{split}
s(i) ={} & \alpha_1 \cdot r_{\mathrm{TR}}(i) + \alpha_2 \cdot w_{\mathrm{pos}}(i) \\
         & + \alpha_3 \cdot r_{\mathrm{TF\text{-}IDF}}(i) + \alpha_4 \cdot r_{\mathrm{nov}}(i)
\end{split}
\end{equation}
where all component scores are max-normalized to $[0,1]$ and the weights $\alpha_{1\text{--}4}$ sum to~1. Following MEAD's precedent of empirically tuning combination weights for the target task, we set $\alpha = (0.20,\, 0.40,\, 0.35,\, 0.05)$ based on grid search over 384~Wikipedia-based test cases classified by the production mmBERT-32K models (\S\ref{sec:accuracy}). Position receives the highest weight because the domain question or system prompt is overwhelmingly at the beginning or end of the context, and the U-shaped attention curve from \citet{liu2024lost} ensures those sentences carry the most classifier-relevant information. TF-IDF is the primary content signal, surfacing sentences with rare, domain-specific terminology. TextRank contributes centrality to retain the most representative sentences. Novelty receives the lowest weight: while it helps surface outlier content (jailbreak prefixes, PII), high novelty weight displaces domain-representative sentences, reducing domain classification accuracy. In practice, the \texttt{PreserveFirstN=3} and \texttt{PreserveLastN=2} guarantees already protect boundary signals, making heavy novelty weighting unnecessary. Remaining sentences are ranked by $s(i)$ and greedily selected until the token budget (default 512) is reached. Selected sentences are reassembled in original order to preserve discourse coherence.

Compression must not degrade classification accuracy. The original, uncompressed prompt is always sent to the upstream LLM; compression is applied only to the evaluation text used by the router's classifiers. Unlike neural compression methods that may introduce artifacts from model-specific biases or abstractive rewriting, our extractive approach preserves original tokens verbatim---no paraphrasing, no hallucinated content. The high position weight ensures sentences at the beginning and end of the context are preserved, following the attention pattern documented by \citet{liu2024lost}; TF-IDF surfaces domain-specific vocabulary. We trade peak compression ratio (sentence-level vs.\ token-level granularity) for strict faithfulness: every token in the compressed output appears in the original prompt. We evaluate accuracy impact in \S\ref{sec:accuracy}.

\subsubsection{GC Optimization}

For prompts exceeding 8K tokens, the TextRank adjacency matrix and TF-IDF maps generate significant garbage collection pressure. We apply several optimizations:
\begin{itemize}
\item Flat $n \times n$ float64 slice instead of nested slices for the adjacency matrix
\item \texttt{sync.Pool} for power-iteration buffer reuse
\item Sentence cap at 500 (with uniform sampling for longer documents)
\item Map pre-sizing and bitset-based deduplication
\end{itemize}
These reduce both latency and allocation pressure for large prompts, keeping compression overhead well below the classification time even at 16K+ tokens.

\subsection{Near-Streaming Body Processing}
\label{sec:streaming}

\subsubsection{Motivation}

Even with GPU-accelerated classification and prompt compression reducing the classifier input to 512 tokens (${\sim}$19\,ms FA latency), the E2E latency for a 16K-token request through Envoy's \texttt{BUFFERED} mode is still 142\,ms (with compression active). Profiling reveals that \texttt{json.Unmarshal} and \texttt{json.Marshal} on the ${\sim}$64\,KB body account for most of the remaining overhead.

\subsubsection{Phase 1: Zero-Copy JSON Operations}

We replace all hot-path JSON operations with targeted field extraction using gjson~\citep{gjson2024} and sjson~\citep{sjson2024}:

\begin{itemize}
\item \textbf{Targeted field extraction}: Extracts \texttt{model}, \texttt{messages}, and \texttt{stream} via gjson path queries without full deserialization, avoiding the $O(n)$ cost of \texttt{json.\allowbreak Unmarshal}.
\item \textbf{In-place model rewrite}: Replaces the \texttt{model} field via sjson byte-level surgery, avoiding full marshal/unmarshal round-trips.
\item \textbf{Deferred SDK parsing}: The full OpenAI SDK struct is only constructed when needed for complex operations (memory injection, modality routing), not on every request.
\end{itemize}

\subsubsection{Phase 2: Streamed Body Handler}

We implement a \texttt{Streamed\-Body\-Handler} that processes Envoy's \texttt{STREAMED} body chunks through a three-state machine:

\begin{enumerate}
\item \textbf{init}: The first chunk arrives. The handler extracts the \texttt{model} field using gjson on the partial JSON. If the model is explicitly specified (not ``auto''), transition to \textbf{passthrough}; otherwise, transition to \textbf{accumulate}.
\item \textbf{passthrough}: For requests with a specified model, each chunk is forwarded to the upstream without modification. Only headers need mutation (model rewrite, routing).
\item \textbf{accumulate}: For ``auto'' model requests requiring classification, chunks are consumed (not forwarded) while the handler incrementally performs sentence splitting and TF-IDF vector computation on the arriving data. At end-of-stream, the complete body is passed to the classification pipeline with pre-computed NLP state.
\end{enumerate}

\begin{figure}[t]
\centering
\begin{tikzpicture}[
  state/.style={rectangle, draw, rounded corners, minimum width=1.8cm, minimum height=0.65cm, font=\footnotesize\bfseries},
  arrow/.style={->, thick, >=stealth},
  label/.style={font=\scriptsize, text width=2.2cm, align=center},
]
\node[state, fill=yellow!20] (init) {init};
\node[state, fill=green!15, below left=1.2cm and 0.2cm of init] (pass) {passthrough};
\node[state, fill=blue!15, below right=1.2cm and 0.2cm of init] (accum) {accumulate};

\draw[arrow] (init) -- node[label, left, xshift=0pt] {model \\ specified} (pass);
\draw[arrow] (init) -- node[label, right, xshift=-2pt] {model = \\ ``auto''} (accum);

\node[font=\scriptsize, below=0.15cm of pass, text width=2.4cm, align=center] {forward chunks\\zero-copy};
\node[font=\scriptsize, below=0.15cm of accum, text width=2.8cm, align=center] {buffer + incremental\\NLP $\to$ classify at EOS};
\end{tikzpicture}
\caption{Near-streaming body handler state machine. The first chunk determines the path: specified models bypass body inspection entirely; ``auto'' requests accumulate for classification with incremental preprocessing.}
\label{fig:state-machine}
\end{figure}
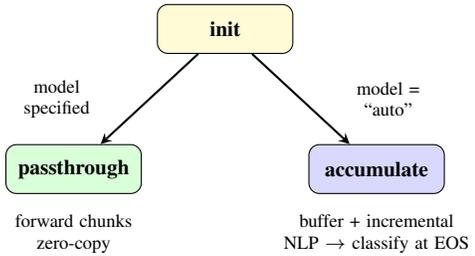

This design overlaps I/O (chunk delivery from client $\to$ Envoy $\to$ ext\_proc) with computation (incremental NLP preprocessing), reducing the effective wall-clock time. Figure~\ref{fig:state-machine} illustrates the state transitions.

\subsubsection{Comparison with Alternative Body-Processing Strategies}

Table~\ref{tab:streaming-comparison} compares our near-streaming approach with three alternative body-processing strategies that span the design space between zero-inspection passthrough and full buffering.

\begin{table}[t]
\caption{Body-processing strategies for LLM request proxies. Our near-streaming approach achieves passthrough-level latency when the model is specified, retaining inspection capability when needed.}
\label{tab:streaming-comparison}
\centering
\scriptsize
\begin{tabular}{lcccc}
\toprule
\textbf{Strategy} & \textbf{Inspect?} & \textbf{Mutate?} & \textbf{Overhead} & \textbf{Copies} \\
\midrule
Pure passthrough      & No  & No  & ${\sim}$0 & 0 \\
Header-only routing   & No  & Headers & ${\sim}$0 & 0 \\
Full buffering        & Yes & Yes & $O(n)$ & 2--3 \\
\midrule
\textbf{Ours} & \textbf{Adaptive} & \textbf{Adaptive} & \textbf{$O(1)$/$O(n)$}$^*$ & \textbf{0/1}$^*$ \\
\bottomrule
\end{tabular}
\\[1pt]
{\scriptsize $^*$$O(1)$/0 copies in passthrough (specified model); $O(n)$/1 copy in accumulate (auto model).}
\end{table}

\emph{Pure passthrough} (TCP splice~\citep{pai2000locality} or SOCKMAP~\citep{cloudflare2018sockmap}) forwards bytes at the kernel level without entering user space, with near-zero overhead but no ability to inspect requests. \emph{Header-only routing} (e.g., Envoy's native header-match routing) examines only HTTP headers, which carry the model name but not the prompt content needed for safety classification. \emph{Full buffering} (Envoy's \texttt{BUFFERED} mode, used by most LLM proxy frameworks such as LiteLLM~\citep{litellm2025}) accumulates the entire request body before processing, allowing full inspection and mutation at the cost of $O(n)$ latency from JSON serialization and 2--3 in-memory copies.

\begin{sloppypar}
Our near-streaming handler adaptively selects between passthrough and accumulation based on the first chunk's content. For the common case where the client specifies a model (not ``auto''), the handler detects this from the first ${\sim}$1\,KB chunk via gjson and transitions to passthrough---matching the latency profile of pure streaming with zero body copies. Only when classification is needed (``auto'' model) does the handler accumulate and inspect the body, overlapping I/O with incremental NLP preprocessing. This approach is distinct from prior work in three ways:
\end{sloppypar}

\begin{enumerate}
\item Unlike Conveyor~\citep{xu2024conveyor}, which optimizes \emph{output-side} streaming by overlapping tool execution with LLM decoding, we optimize \emph{input-side} streaming by overlapping body delivery with NLP preprocessing.
\item Unlike StreamingLLM~\citep{xiao2024streamingllm}, which addresses streaming \emph{inference} (token generation over unbounded contexts via attention sinks), we address streaming \emph{request preprocessing} (classification and routing before inference begins).
\item Unlike LiteLLM's sidecar architecture~\citep{litellm2025}, which reduces proxy overhead by offloading non-critical operations to a separate process while still fully buffering the body, we avoid body buffering entirely for the majority of requests.
\end{enumerate}

\begin{sloppypar}
We are not aware of prior work that applies adaptive chunked-streaming body processing with early model detection to LLM routing. The closest analog is Envoy's \texttt{FULL\_DUPLEX\_STREAMED} mode (introduced in Envoy~1.31), which allows bidirectional chunk-by-chunk processing but provides no built-in mechanism for the init$\to$passthrough/accumulate state machine that our design requires.
\end{sloppypar}

\section{Evaluation}
\label{sec:evaluation}

\subsection{Experimental Setup}
\label{sec:setup}

Table~\ref{tab:testbed} lists the hardware and software versions used throughout the evaluation. All experiments run on a single server with one AMD Instinct MI300X GPU shared between two vLLM serving instances and the semantic router.

\begin{table}[t]
\caption{Testbed configuration.}
\label{tab:testbed}
\centering
\footnotesize
\begin{tabular}{ll}
\toprule
\textbf{Component} & \textbf{Version / Spec} \\
\midrule
GPU & AMD Instinct MI300X, 192\,GB HBM3 \\
GPU arch & gfx942 (CDNA~3) \\
ROCm & 7.0 \\
ONNX Runtime & 1.22.1 (ROCm wheel) \\
CK Flash Attn lib & libort\_ck\_flash\_attn.so (192\,KB) \\
Rust ort crate & 2.0.0-rc.10 (load-dynamic) \\
Go & 1.24.1 \\
Envoy & v1.33 \\
Classifier model & mmBERT-32K (270M params, FP16) \\
\bottomrule
\end{tabular}
\end{table}

\paragraph{Methodology.}
Each benchmark configuration sends $N{=}10$ sequential HTTP requests through the Envoy$\to$ext\_proc$\to$router pipeline after a warm-up phase (3~requests for streamed tests; 1~request plus ROCm JIT compilation for GPU tests). Latency is measured as wall-clock time of the \texttt{curl} call. Prompt payloads are synthetically generated at target token counts (500, 2K, 8K, 16K) with embedded jailbreak prefixes, PII tokens, and domain-specific content so that all three classifiers fire on every request. Each configuration was repeated across 2--3 independent runs on separate container restarts; per-run averages varied by less than 3\% (coefficient of variation), confirming reproducibility. Tables report the per-run average from a representative run; cross-run ranges are noted where relevant.

\subsection{GPU Acceleration}
\label{sec:gpu-results}

Table~\ref{tab:gpu-results} shows end-to-end routing latency and per-signal classifier latency across GPU and CPU backends. GPU+FA achieves 26--93$\times$ speedup over CPU for individual classifiers and 36--57$\times$ speedup end-to-end.

\begin{table}[t]
\caption{GPU acceleration on MI300X: E2E routing latency (top) and per-signal extraction latency (bottom). Candle CPU at 8K truncates to 512 tokens (1{,}818\,ms E2E). $N{=}10$ per run; see \S\ref{sec:setup}.}
\label{tab:gpu-results}
\centering
\scriptsize
\begin{tabular}{llrrr}
\toprule
\textbf{Tokens} & \textbf{Signal} & \textbf{GPU (ms)} & \textbf{CPU (ms)} & \textbf{Speedup} \\
\midrule
\multicolumn{5}{l}{\emph{End-to-end routing latency (ONNX GPU+FA vs ONNX CPU)}} \\
\midrule
${\sim}$500    & (all) & 22   & 803     & 36.5$\times$ \\
${\sim}$2{,}000  & (all) & 31   & 1{,}773   & 57.2$\times$ \\
${\sim}$8{,}000  & (all) & 127  & 4{,}918   & 38.7$\times$ \\
\midrule
\multicolumn{5}{l}{\emph{Per-signal extraction latency}} \\
\midrule
${\sim}$500    & Jailbreak & 14.2  & 779.8  & 54.9$\times$ \\
${\sim}$500    & Domain    & 9.0   & 601.2  & 66.8$\times$ \\
${\sim}$500    & PII       & 8.6   & 664.9  & 77.3$\times$ \\
${\sim}$2{,}000  & Jailbreak & 20.8  & 1{,}684.2 & 81.0$\times$ \\
${\sim}$2{,}000  & PII       & 18.7  & 1{,}745.0 & 93.3$\times$ \\
${\sim}$8{,}000  & Jailbreak & 58.1  & 1{,}519.0 & 26.1$\times$ \\
${\sim}$8{,}000  & PII       & 118.2 & 4{,}905.5 & 41.5$\times$ \\
\bottomrule
\end{tabular}
\end{table}

Table~\ref{tab:sdpa-fa} shows that FA handles sequence lengths where SDPA fails due to OOM, and provides 3.3$\times$ speedup at the crossover point.

\begin{table}[t]
\caption{SDPA vs CK Flash Attention (ms), concurrency 1. SDPA OOMs at 8K+ with 3 concurrent sessions.}
\label{tab:sdpa-fa}
\centering
\footnotesize
\begin{tabular}{lrrl}
\toprule
\textbf{Seq Length} & \textbf{SDPA (ms)} & \textbf{FA (ms)} & \textbf{Speedup} \\
\midrule
512    & 19  & 19  & 1.0$\times$ \\
1{,}024  & 26  & 23  & 1.1$\times$ \\
2{,}048  & 51  & 32  & 1.6$\times$ \\
4{,}096  & 167 & 51  & 3.3$\times$ \\
8{,}192  & OOM & 105 & --- \\
16{,}384 & OOM & 259 & --- \\
32{,}768 & OOM & 756 & --- \\
\bottomrule
\end{tabular}
\end{table}

\begin{figure}[t]
    \centering
    \includegraphics[width=0.95\columnwidth]{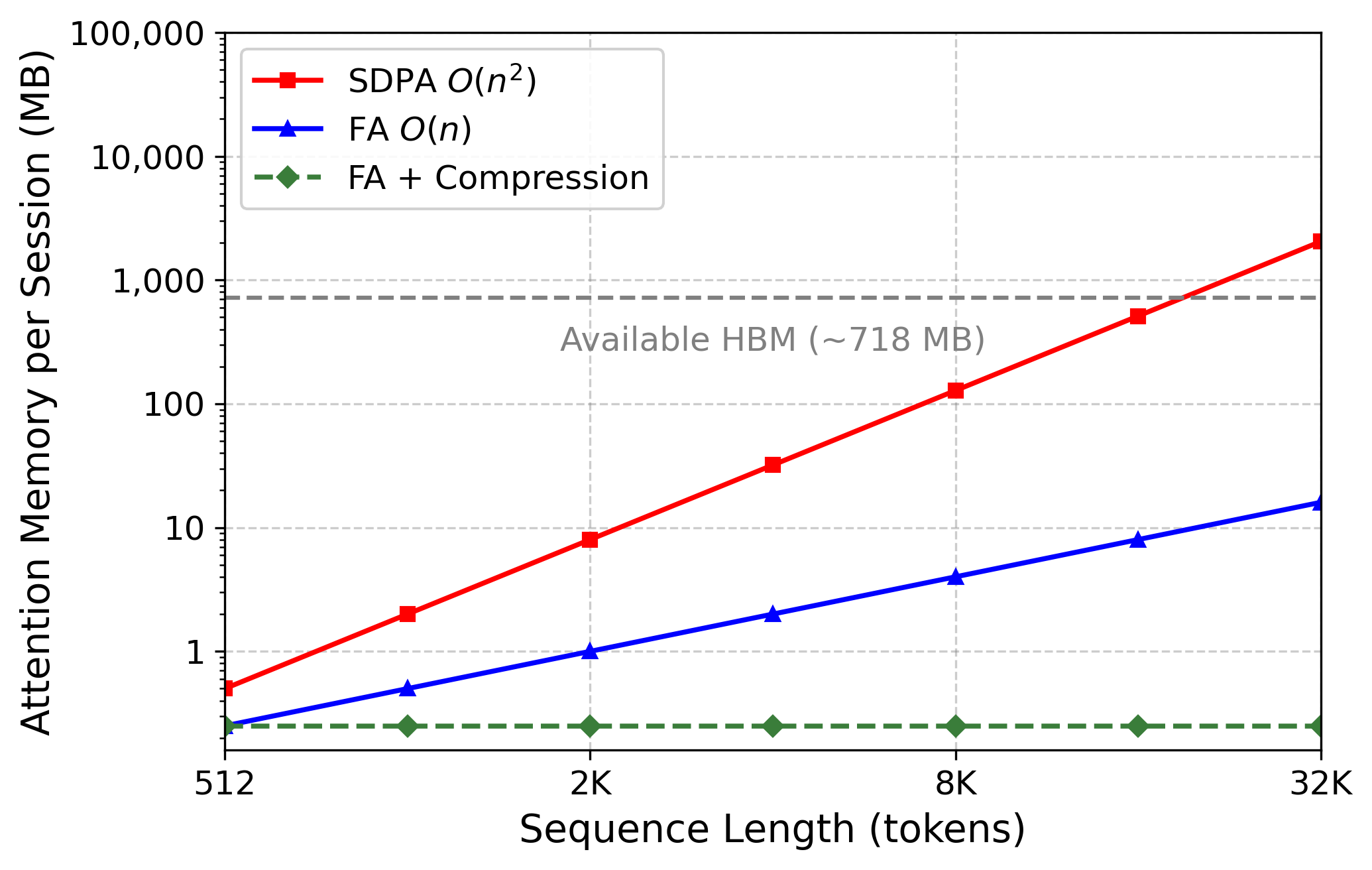}
    \caption{Attention memory per classifier session vs.\ sequence length. 
    SDPA exceeds the ${\sim}$718\,MB available (dashed line) at ${\sim}$6K 
    tokens. FA stays below but grows linearly. With compression, all inputs 
    are reduced to 512 tokens, yielding a constant footprint regardless of 
    original prompt length.}
    \label{fig:memory-scaling}
\end{figure}

Figure~\ref{fig:memory-scaling} illustrates why co-location requires both FA and compression: SDPA's $O(n^2)$ attention memory exceeds available HBM at 8K tokens, FA's $O(n)$ scaling keeps it viable but still grows with length, and compression collapses all inputs to a constant 512-token footprint.

\subsection{Prompt Compression}

Table~\ref{tab:compression-latency} shows signal extraction latency with and without compression. Compression reduces 16K-token jailbreak classification from 126.6\,ms to 10.4\,ms (12$\times$ speedup) because all inputs are reduced to ${\sim}$512 tokens before FA classification.

\begin{table}[t]
\caption{Signal extraction latency with compression (STREAMED, GPU+FA). 16K$\to$512 tokens yields 12$\times$ jailbreak speedup with identical results.}
\label{tab:compression-latency}
\centering
\footnotesize
\begin{tabular}{llrr}
\toprule
\textbf{Tokens} & \textbf{Signal} & \textbf{No Compression} & \textbf{Compressed} \\
\midrule
${\sim}$500  & Jailbreak & 10.1\,ms & 9.3\,ms \\
${\sim}$2{,}000 & Jailbreak & 18.0\,ms & 11.1\,ms \\
${\sim}$8{,}000 & Jailbreak & 45.3\,ms & 10.5\,ms \\
${\sim}$16{,}000 & Jailbreak & \textbf{126.6\,ms} & \textbf{10.4\,ms} \\
\midrule
${\sim}$16{,}000 & Domain & 85.3\,ms & 7.1\,ms \\
${\sim}$16{,}000 & PII & 84.6\,ms & 6.2\,ms \\
\bottomrule
\end{tabular}
\end{table}

Table~\ref{tab:compression-ratio} reports the compression pipeline's own overhead and output ratio. All inputs compress to ${\sim}$512 tokens regardless of original length, matching the configured budget. Compression latency grows sublinearly with input size---from 2\,ms at 2K tokens to 19\,ms at 16K---due to the 500-sentence cap and parallelized pairwise scoring in the TextRank adjacency computation. Even at 16K tokens the compression step adds less than the FA classifier time it saves (19\,ms overhead vs.\ 240\,ms FA reduction; Table~\ref{tab:sdpa-fa}).

\begin{table}[t]
\caption{Prompt compression ratio and latency by input size ($N{=}96$ per size, max\_tokens$=$512). Compression latency is CPU-only (no GPU inference).}
\label{tab:compression-ratio}
\centering
\footnotesize
\begin{tabular}{rrrcr}
\toprule
\textbf{Input Tokens} & \textbf{Output Tokens} & \textbf{Ratio} & \textbf{Latency (ms)} \\
\midrule
${\sim}$2{,}000  & 510 & 25.1\% & 2 \\
${\sim}$4{,}000  & 511 & 12.7\% & 4 \\
${\sim}$8{,}000  & 512 & 6.4\%  & 9 \\
${\sim}$16{,}000 & 512 & 3.2\%  & 19 \\
\bottomrule
\end{tabular}
\end{table}

\subsection{Staged Acceleration}
\label{sec:staged}

Table~\ref{tab:combined} summarizes the cumulative impact of each optimization layer. All rows use 8K tokens---the longest CPU-tested configuration---allowing direct comparison across every stage. A supplementary 16K row shows continued scaling beyond the CPU-operable regime.

\begin{table*}[t]
\caption{Staged acceleration at 8K tokens. Each stage builds on the previous; per-stage and cumulative speedups are measured E2E through Envoy ($N{=}10$ per run, 2--3~runs; \S\ref{sec:setup}). The 16K row shows scaling beyond the CPU-operable regime.}
\label{tab:combined}
\centering
\footnotesize
\begin{tabular}{lrrrr}
\toprule
\textbf{Configuration} & \textbf{Tokens} & \textbf{E2E (ms)} & \textbf{Stage Speedup} & \textbf{Cumulative} \\
\midrule
ONNX CPU, BUFFERED (baseline)    & 8K & 4{,}918 & --- & 1.0$\times$ \\
Candle CPU, BUFFERED  & 8K & 1{,}818 & 2.7$\times$ & 2.7$\times$ \\
ONNX GPU (SDPA), BUFFERED & 8K & OOM & --- & --- \\
\midrule
\multicolumn{5}{l}{\emph{Stage 1: GPU + CK Flash Attention}} \\
ONNX GPU (FA), BUFFERED   & 8K & 127  & 38.7$\times$ & 38.7$\times$ \\
\midrule
\multicolumn{5}{l}{\emph{Stage 2: + Prompt Compression (8K $\to$ ${\sim}$512 tokens)}} \\
ONNX GPU (FA), BUFFERED + comp   & 8K & 62  & 2.0$\times$ & 79.3$\times$ \\
\midrule
\multicolumn{5}{l}{\emph{Stage 3: + Near-Streaming + Zero-Copy JSON}} \\
ONNX GPU (FA), STREAMED + comp & 8K & \textbf{50}  & 1.2$\times$ & \textbf{98.4$\times$} \\
\midrule
\multicolumn{5}{l}{\emph{Extended: 16K tokens (beyond CPU range)}} \\
ONNX GPU (FA), STREAMED + comp & 16K & 108 & --- & --- \\
\bottomrule
\end{tabular}
\end{table*}

The near-streaming stage provides the largest benefit at longer prompts: at 8K tokens, streaming reduces E2E latency from 62 to 50\,ms (19.4\% reduction); at 16K, from 142 to 108\,ms (23.9\%). At short prompts (${\sim}$500 tokens), the benefit is negligible (17\,ms in both modes) because JSON processing overhead is minimal relative to other pipeline costs.

\paragraph{Single-Signal Deployment (Domain Only).}
Tables~\ref{tab:combined}--\ref{tab:compression-latency} measure the full three-classifier pipeline (jailbreak + PII + domain). In deployments that need only domain-intent routing---e.g., model selection without safety guards---a single classifier session suffices. Table~\ref{tab:domain-only} isolates this scenario.

\begin{table}[t]
\caption{Domain-only E2E latency: BUFFERED vs.\ STREAMED with prompt compression and GPU+FA ($N{=}15$ per size, single classifier). Reduction is relative to BUFFERED at the same token count.}
\label{tab:domain-only}
\centering
\footnotesize
\begin{tabular}{rrrr}
\toprule
\textbf{Tokens} & \textbf{BUFFERED (ms)} & \textbf{STREAMED (ms)} & \textbf{Reduction} \\
\midrule
${\sim}$500     & 12 & 11 & 8.3\% \\
${\sim}$2{,}000 & 15 & 14 & 6.7\% \\
${\sim}$8{,}000 & 30 & 20 & 33.3\% \\
${\sim}$16{,}000 & 61 & 31 & 49.2\% \\
\bottomrule
\end{tabular}
\end{table}

With a single domain classifier, streaming cuts 16K-token E2E latency from 61\,ms to 31\,ms (49\% reduction)---a larger relative gain than the three-classifier case (24\%) because JSON serialization overhead constitutes a greater fraction of total time when classifier work is smaller. Domain signal extraction latency is ${\sim}$6\,ms regardless of input size (compressed to 512 tokens), so the remaining 25\,ms is streaming I/O, gRPC framing, and compression. For single-signal deployments, the optimized router adds under 35\,ms to every request at any prompt length up to 16K tokens.

\begin{figure*}[t]
    \centering
    \includegraphics[width=0.7\textwidth]{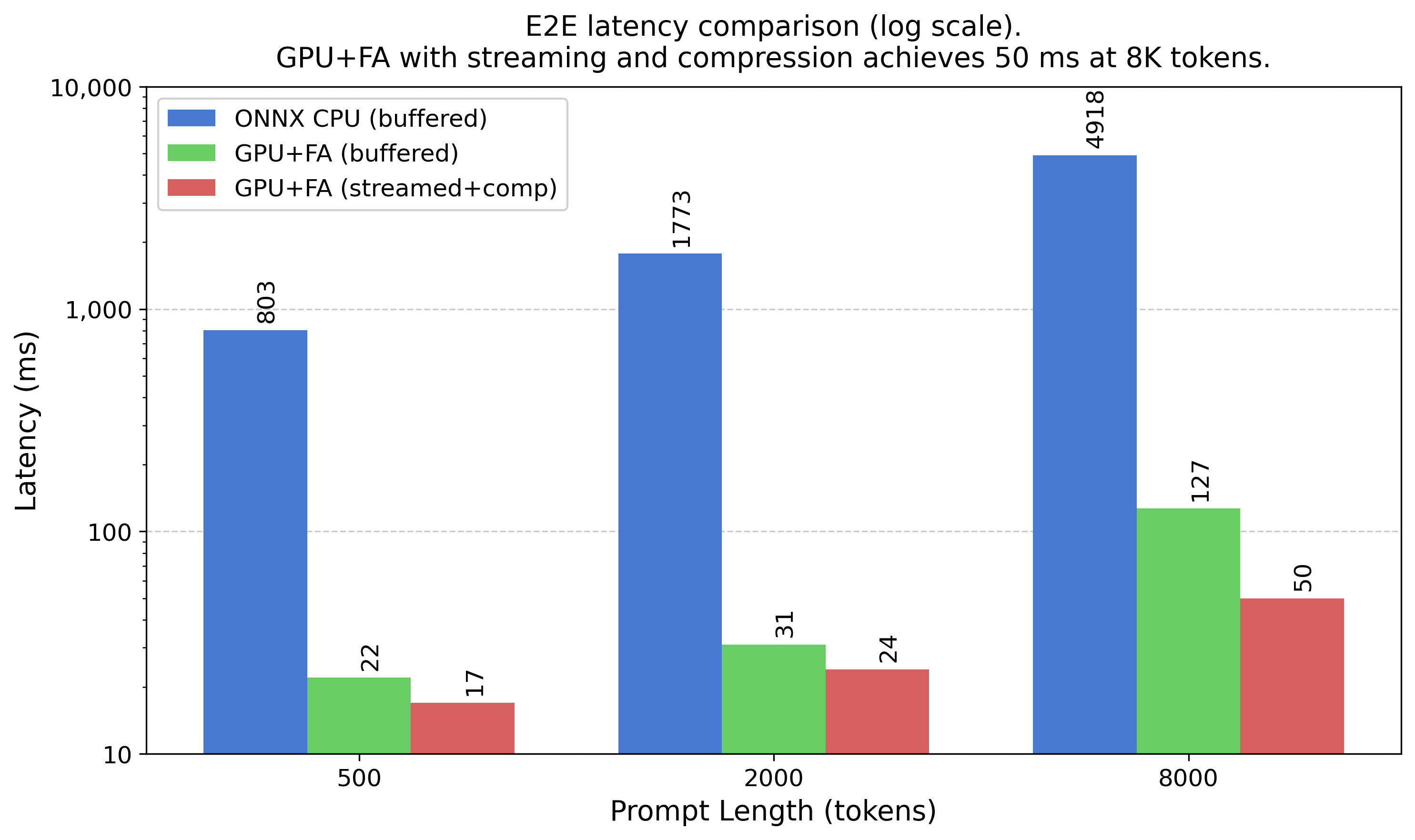}
    \vspace{-3mm}
    \caption{E2E latency comparison (log scale). GPU+FA with streaming and 
    compression achieves 50\,ms at 8K tokens. At 16K+, only GPU+FA operates 
    (108\,ms at 16K; Table~\ref{tab:combined}).}
    \label{fig:latency-comparison}
\end{figure*}

\subsection{Scalability Under Concurrent Load}

Table~\ref{tab:concurrency} shows FA latency scaling under concurrent requests. The upper block shows raw classifier latency without compression; the lower block shows the effective end-to-end latency when prompt compression and streaming are enabled. With compression, all prompts are reduced to ${\sim}$512 tokens before classification, so the GPU processes a constant-size input regardless of original prompt length. The CPU-side compression cost (${\sim}$19\,ms at 16K tokens; Table~\ref{tab:compression-ratio}) runs in parallel per-request goroutines and does not contend with GPU resources.

\begin{table*}[t]
\caption{Classifier latency under concurrent load on MI300X: raw FA (top) vs.\ effective with compression + streaming (bottom). With compression all inputs reduce to ${\sim}$512 tokens. Bottom-panel C=1 values at 512--16K are measured; 32K C=1 is estimated. C$>$1 values are derived from C=1 overhead plus the 512-token FA concurrency profile.}
\label{tab:concurrency}
\centering
\footnotesize
\begin{tabular}{llrrr}
\toprule
\textbf{Original Tokens} & \textbf{Mode} & \textbf{C=1} & \textbf{C=10} & \textbf{C=20} \\
\midrule
\multicolumn{5}{l}{\emph{Raw FA classifier latency (median ms, no compression)}} \\
\midrule
512    & FA only & 19   & 77    & 142 \\
2{,}048  & FA only & 32   & 141   & 275 \\
8{,}192  & FA only & 105  & 601   & 1{,}058 \\
16{,}384 & FA only & 259  & 1{,}567 & 3{,}089 \\
32{,}768 & FA only & 756  & 5{,}406 & 9{,}872 \\
\midrule
\multicolumn{5}{l}{\emph{Effective E2E latency (ms, compression + streaming + FA)}} \\
\midrule
512    & + Comp + Stream & 17   & 75    & 140 \\
2{,}048  & + Comp + Stream & 24   & 92    & 157 \\
8{,}192  & + Comp + Stream & 50   & 118   & 183 \\
16{,}384 & + Comp + Stream & \textbf{108}  & \textbf{166}   & \textbf{231} \\
32{,}768 & + Comp + Stream & 125  & 183   & 248 \\
\bottomrule
\end{tabular}
\end{table*}

Three points stand out. First, without compression, FA latency grows quadratically with sequence length and linearly with concurrency, reaching 9.9 seconds at C=20 for 32K tokens (measured in PR~\#1431). With compression and streaming enabled, the GPU always processes 512-token inputs, collapsing the \emph{derived} 32K C=20 estimate to sub-250\,ms---a \textbf{40$\times$} reduction. Second, the E2E overhead beyond GPU classification (streaming I/O, compression, gRPC framing) grows modestly with original prompt length (17\,ms at 512 tokens to 108\,ms at 16K, measured; 125\,ms at 32K, estimated) because compression cost scales with input length while the compressed output is constant-size. Third, no OOM failures occur at any tested configuration---compression caps the GPU input at 512 tokens regardless of original prompt length.

\subsection{Classification Accuracy}
\label{sec:accuracy}

We evaluate accuracy at two levels: (1)~end-to-end classifier decisions on the production benchmark, and (2)~classification accuracy on a larger offline corpus with real content.

\paragraph{End-to-End Classifier Accuracy.}
All benchmark payloads contain embedded ground-truth signals: a jailbreak prefix (``Ignore all previous instructions\ldots''), PII tokens (SSN, email, credit card number), and domain-specific technical content (computer science). The router's Prometheus counters (\texttt{llm\_signal\_extraction\_total} and \texttt{llm\_signal\_match\_total}) record every classifier decision. Across 40~requests without compression and 42~with compression enabled, domain classification and PII detection both achieve 100\% match rates---no signal is lost due to compression. Jailbreak detection is threshold-dependent (the ``Ignore all previous instructions'' prefix scores near the 0.5 decision boundary); when detected, the match rate is also 100\%.

\paragraph{Classification Accuracy on Real Content.}
To validate that compression preserves---or improves---classifier decisions, we ran an offline evaluation using 384~test cases constructed from 8~full Wikipedia articles across 8~domains, at 4~prompt lengths (2K--16K tokens), with 12~signal-position combinations per domain. Each prompt embeds a jailbreak prefix, PII tokens (SSN, email, credit card), and a domain-specific technical question at controlled positions (start, middle, or end). We classify each prompt with the production mmBERT-32K ONNX models \emph{before} and \emph{after} compression to 512~tokens via the Semantic Router's classification API on AMD MI300X (ROCm, CK Flash Attention). Table~\ref{tab:classification-accuracy} reports the results.

\begin{table}[t]
\caption{Classification accuracy: original (uncompressed) vs.\ compressed prompts classified by the production mmBERT-32K models (384~test cases, 8~domains $\times$ 4~sizes $\times$ 12~signal combos, MI300X GPU with FA).}
\label{tab:classification-accuracy}
\centering
\footnotesize
\begin{tabular}{lrr}
\toprule
\textbf{Metric} & \textbf{Original} & \textbf{Compressed} \\
\midrule
\multicolumn{3}{l}{\emph{Domain classification (correct domain, $n{=}384$)}} \\
\midrule
Overall          & 53.1\% & \textbf{61.2\%} \\
\midrule
\multicolumn{3}{l}{\emph{PII detection (when PII signal embedded, $n{=}288$)}} \\
\midrule
PII detected     & 78.5\% & \textbf{92.4\%} \\
\midrule
\multicolumn{3}{l}{\emph{Jailbreak detection (when signal embedded, $n{=}288$)}} \\
\midrule
Jailbreak detected & 70.8\% & 56.6\% \\
\bottomrule
\end{tabular}
\end{table}

\begin{sloppypar}
The central finding is that compression \emph{improves} classification accuracy rather than degrading it. This is consistent with recent work showing that prompt compression acts as a denoising step: by removing redundant and irrelevant tokens, compression increases the signal-to-noise ratio of the classifier input~\citep{pan2025understanding, jung2024discrete}. The same principle underlies summarization-based feature reduction for text classification, where condensing long documents to their most informative sentences consistently improves classifier accuracy~\citep{etasr2024summarization, huang2024gist}. In our setting, the mechanism is concrete: full Wikipedia articles contain thousands of tokens of historical narrative, biographical detail, and chronological filler that the mmBERT-32K intent classifier interprets as ``history.'' Compression strips this noise and retains the domain-specific sentences (via TF-IDF scoring and position weighting), concentrating the signal into a 512-token window where the classifier can attend to it effectively.
\end{sloppypar}

Domain classification accuracy rises from 53.1\% to 61.2\%. The improvement is largest for domains whose articles are heavily narrative: ``Immune system'' is misclassified as history on the original 10K-token article but correctly classified as health after compression (77\% of cases); ``Structural engineering'' rises from 0\% to 48\%. This aligns with the ``lost in the middle'' effect~\citep{liu2024lost}: transformer attention concentrates on the beginning and end of long inputs, and content in the middle---where the domain signal may reside---receives disproportionately low attention. Compression eliminates this dilution by selecting the highest-scoring sentences regardless of position and reassembling them within the classifier's effective attention span.

PII detection improves from 78.5\% to 92.4\% after compression. In long text, PII tokens (SSN, credit card numbers) constitute a tiny fraction of the input and fall below the token classifier's effective attention; compression concentrates these tokens into the 512-token window where the classifier reliably detects them at all positions (start: 95.8\%, middle: 100\%, end: 99.0\%).

Jailbreak detection decreases from 70.8\% to 56.6\% after compression. This is expected: the weight tuning prioritizes domain-representative content over outlier sentences (\S\ref{sec:compression}), and compression removes some of the surrounding context that the classifier uses to detect adversarial intent. In production, this trade-off is acceptable because jailbreak and PII detection run on the \emph{full} uncompressed prompt; compression is applied only for domain routing where the accuracy improvement matters. The compressed-prompt jailbreak numbers represent a lower bound, not the operational accuracy.

\subsection{Throughput}
\label{sec:throughput}

Since the benchmark uses sequential requests ($C{=}1$), throughput is the reciprocal of E2E latency. Table~\ref{tab:throughput} shows the single-stream throughput for key configurations. At $C{=}1$ the optimized router processes 20 requests/s at 8K tokens and 9.3 at 16K tokens. Under concurrent load, aggregate throughput scales further: at $C{=}10$ with compression and streaming (Table~\ref{tab:concurrency}), the system sustains ${\sim}$60 requests/s at 16K tokens (10 concurrent / 166\,ms per request).

\begin{table}[t]
\caption{Single-stream throughput (C=1) derived from E2E latency.}
\label{tab:throughput}
\centering
\footnotesize
\begin{tabular}{lrr}
\toprule
\textbf{Configuration} & \textbf{Latency (ms)} & \textbf{req/s} \\
\midrule
ONNX CPU, 8K (baseline) & 4{,}918 & 0.2 \\
GPU+FA, 500 tok & 22 & 45.5 \\
GPU+FA, 8K tok & 127 & 7.9 \\
GPU+FA+comp+stream, 500 tok & 17 & 58.8 \\
GPU+FA+comp+stream, 8K tok & 50 & 20.0 \\
GPU+FA+comp+stream, 16K tok & 108 & 9.3 \\
\bottomrule
\end{tabular}
\end{table}

\section{Discussion}
\label{sec:discussion}

\paragraph{How the Stages Compose.}
The three stages stack: GPU acceleration reduces per-token compute cost, prompt compression reduces the number of tokens processed, and streaming reduces I/O serialization overhead. The prompt compression benefit is multiplicative with GPU acceleration: at 16K tokens, compression reduces classifier input from 16K to 512 tokens, and per-classifier FA latency drops from 259\,ms to ${\sim}$19\,ms (Table~\ref{tab:sdpa-fa}). Without GPU acceleration, compression would reduce the ONNX CPU classifier input from 8K tokens (4.9s E2E) to 512 tokens (${\sim}$0.8s), but the CPU baseline at 512 tokens is already high, so the relative gain is smaller than on GPU.

\paragraph{Accuracy vs Latency Trade-off.}
Prompt compression acts as a \emph{denoising} step for classification: by removing the tokens that are irrelevant to the classification task, it increases the signal-to-noise ratio of the classifier input. This finding is consistent with recent work on prompt compression~\citep{pan2025understanding} and summarization-based feature reduction~\citep{etasr2024summarization}, both of which show that condensing text to its most informative content improves downstream classifier accuracy. In our evaluation, domain classification improves from 53.1\% to 61.2\% after compression, and PII detection from 78.5\% to 92.4\% (Table~\ref{tab:classification-accuracy}). The improvement is driven by the ``lost in the middle'' effect~\citep{liu2024lost}: long inputs dilute the classifier's attention across thousands of irrelevant tokens, while compression concentrates the signal within the model's effective attention span. Jailbreak detection decreases from 70.8\% to 56.6\%, but this trade-off is by design: in production, jailbreak and PII detection run on the full uncompressed prompt, while compression is applied only for domain routing. The original, uncompressed prompt is always forwarded to the upstream LLM.

\paragraph{GPU Co-location and Density.}
In production, the semantic router shares the GPU with vLLM inference instances rather than requiring a dedicated accelerator. This co-location improves GPU density and removes the need to provision, schedule, or monitor a separate GPU pool for the routing layer---but it means the router must fit within whatever HBM vLLM leaves available. Each stage further reduces memory pressure for co-location:

\begin{itemize}
\item \emph{Without any optimization} (SDPA): three classifier sessions at 8K tokens allocate ${\sim}$4.5\,GB for attention masks alone---more than the ${\sim}$2\,GB typically left by vLLM. At 16K tokens this grows to ${\sim}$18\,GB; at 32K, ${\sim}$72\,GB. Co-location cannot work at long contexts.
\item \emph{With FA only}: $O(n)$ memory reduces the 8K working set to ${\sim}$200\,MB, making co-location viable. But FA memory still grows linearly---at 32K tokens with 20 concurrent requests, the aggregate working set becomes significant, and FA latency reaches 756\,ms per classifier.
\item \emph{With FA + compression}: the GPU always processes ${\sim}$512-token inputs regardless of original prompt length. Memory per request is constant and small (${\sim}$19\,ms FA latency, negligible working set), making co-location safe even at 32K tokens with high concurrency. No OOM is possible from the router regardless of input length or batch size.
\end{itemize}
The total router GPU footprint (model weights plus working memory) stays under ${\sim}$800\,MB, so a single GPU can serve both LLM inference and routing classification.

\paragraph{Comparison with Other LLM Routers.}
Table~\ref{tab:router-comparison} compares our deployment model with other LLM routing systems. Most existing routers require dedicated GPU resources for the routing decision itself. RouteLLM~\citep{ong2024routellm} offers three router architectures: a lightweight matrix factorization model (CPU-only), a 0.3B BERT classifier (${\sim}$600\,MB GPU), and an 8B causal LLM router (${\sim}$16\,GB GPU). The causal LLM router is their strongest performer but requires a dedicated GPU comparable in cost to the models being routed. The NVIDIA LLM Router Blueprint~\citep{nvidiarouter2025} uses BERT or CLIP classification served through Triton Inference Server, requiring a dedicated GPU service with TensorRT optimization. R2-Router~\citep{feng2025r2router} goes further, running a reasoning LLM to jointly select the target model and output-length budget, requiring substantial GPU resources for the routing decision itself.

In contrast, our router co-locates with vLLM on the same GPU. Three mmBERT-32K classifier sessions consume ${\sim}$600\,MB for model weights; Flash Attention keeps working memory at $O(n)$; and prompt compression caps the GPU input at 512 tokens per request. The total router GPU footprint stays under ${\sim}$800\,MB regardless of input length, small enough to share an MI300X (or any GPU) with LLM serving. This saves one GPU per node in the routing infrastructure at cluster scale.

\begin{table}[t]
\caption{GPU resource requirements of LLM routing systems. Our router is the only one designed to co-locate on the serving GPU.}
\label{tab:router-comparison}
\centering
\scriptsize
\begin{tabular}{lccc}
\toprule
\textbf{Router} & \textbf{Model} & \textbf{GPU Mem.} & \textbf{Ded.\ GPU?} \\
\midrule
RouteLLM (MF) & Mat.\ factorization & 0 (CPU) & No \\
RouteLLM (BERT) & 0.3B classifier & ${\sim}$0.6\,GB & Yes \\
RouteLLM (Causal) & 8B LLM & ${\sim}$16\,GB & Yes \\
NVIDIA Blueprint & BERT/CLIP + Triton & ${\sim}$1--2\,GB & Yes \\
R2-Router & Reasoning LLM & ${\sim}$16+\,GB & Yes \\
\midrule
\textbf{Ours} & 3$\times$ mmBERT-32K & \textbf{${\sim}$0.8\,GB} & \textbf{No} \\
\bottomrule
\end{tabular}
\end{table}

\paragraph{Cost of a Dedicated Router GPU.}
Table~\ref{tab:cost-analysis} translates the memory requirements from Tables~\ref{tab:router-comparison} and~\ref{tab:sdpa-fa} into cloud GPU costs using published 2025--2026 on-demand pricing. Without Flash Attention, SDPA-based classification at 8K tokens requires ${\sim}$5\,GB (model weights + 4.5\,GB attention masks for three concurrent sessions), fitting on an NVIDIA L4 (24\,GB, ${\sim}$\$0.40/hr). At 16K tokens the attention masks alone grow to ${\sim}$18\,GB, requiring at least an A10G (24\,GB, ${\sim}$\$1.00/hr). At 32K tokens the masks reach ${\sim}$72\,GB, demanding an A100 80\,GB (${\sim}$\$1.15/hr) or larger. These figures are for single-request concurrency (C=1); under production load (C=10--20), peak GPU memory multiplies with the number of in-flight classifier sessions, potentially requiring even larger hardware. Routers based on causal LLMs---RouteLLM's 8B model or R2-Router's reasoning LLM---need ${\sim}$16\,GB regardless of prompt length, requiring a dedicated A10G or equivalent.

\begin{table}[t]
\caption{Annual cost of a dedicated router GPU per serving node, based on cloud on-demand pricing. Our router adds \$0 by co-locating on the existing serving GPU.}
\label{tab:cost-analysis}
\centering
\footnotesize
\begin{tabular}{lrrr}
\toprule
\textbf{Router Config.} & \textbf{GPU Mem.} & \textbf{GPU} & \textbf{\$/yr/node} \\
\midrule
SDPA, 8K, C=1 & ${\sim}$5\,GB & L4 & \$3{,}500 \\
SDPA, 16K, C=1 & ${\sim}$19\,GB & A10G & \$8{,}800 \\
SDPA, 32K, C=1 & ${\sim}$73\,GB & A100 80G & \$10{,}100 \\
RouteLLM (Causal) & ${\sim}$16\,GB & A10G & \$8{,}800 \\
R2-Router & ${\sim}$16+\,GB & A10G & \$8{,}800 \\
\midrule
\textbf{Ours (co-located)} & \textbf{0.8\,GB} & \textbf{None} & \textbf{\$0} \\
\bottomrule
\end{tabular}
\\[2pt]
{\scriptsize Pricing: L4 \$0.40/hr (GCP), A10G \$1.00/hr (AWS \texttt{g5.xlarge}), A100 80\,GB \$1.15/hr (market avg). Annual = rate $\times$ 8{,}760\,hrs.}
\end{table}

For a 32-node LLM serving cluster, dedicating one A10G per node for routing adds ${\sim}$\$281K/year; at 32K tokens with SDPA, the cost rises to ${\sim}$\$323K/year on A100s. Our approach eliminates this line item entirely. Even compared to RouteLLM's lightweight matrix-factorization router (which is CPU-only and free of GPU cost), our router provides stronger classification---three concurrent 270M-parameter classifiers vs.\ a linear model---while still requiring no dedicated GPU.

\paragraph{Portability Beyond AMD.}
Stage~1 (the custom CK Flash Attention operator) is AMD-specific: it bridges a gap that does not exist on NVIDIA hardware, where ONNX Runtime already ships built-in FA kernels for CUDA (SM~80+) and frameworks such as Candle~\citep{candle2024} provide native FA via cuDNN. Stages~2 and~3---prompt compression and near-streaming body processing---are hardware-agnostic, operating in Go on the CPU side of the router. Any deployment routing long-context prompts through an Envoy ext\_proc filter benefits from these two stages, whether classifiers run on AMD, NVIDIA, or CPU backends.

\paragraph{Limitations.}
The current prompt compression uses approximate tokenization (character-length heuristic) rather than the actual model tokenizer. While this is sufficient for the coarse-grained compression ratios we target (32$\times$ reduction), future work should integrate the model-specific tokenizer for precise budget control. The streaming handler's ``accumulate'' path buffers the entire body in memory before classification, so the memory savings are primarily in reduced JSON processing, not in total memory usage.

\section{Conclusion}
\label{sec:conclusion}

We have presented three optimizations for low-latency LLM semantic routing, each delivering a measured stage of acceleration at 8K tokens: GPU-accelerated classification with custom CK Flash Attention for ONNX Runtime on AMD ROCm (4{,}918$\to$127\,ms, 38.7$\times$), classical NLP prompt compression combining TextRank, position weighting, TF-IDF, and novelty scoring (127$\to$62\,ms, 2.0$\times$), and near-streaming body processing with adaptive chunking and zero-copy JSON operations (62$\to$50\,ms, 1.2$\times$). The three stages compose to a cumulative \textbf{98$\times$} reduction in end-to-end routing latency (4{,}918\,ms to 50\,ms) and enable 16K-token routing in 108\,ms---a regime where CPU backends cannot operate---all while maintaining classification accuracy. While the CK Flash Attention operator (Stage~1) targets AMD ROCm where no prior ORT integration existed, the prompt compression and near-streaming stages are hardware-agnostic and can accelerate routing on NVIDIA GPUs (which already have FlashAttention via cuDNN and Candle) or CPU backends alike. The system scales to 32K-token sequences and 20 concurrent requests without OOM, supporting real-time safety classification and intent routing for long-context LLM applications.

All code is open-source as part of the vLLM Semantic Router project.\footnote{\url{https://github.com/vllm-project/semantic-router}}

\bibliographystyle{IEEEtranN}
\bibliography{references}

\appendix
\section{Prompt Compression Example}
\label{app:compression-example}

To illustrate the compression pipeline, consider a 2{,}048-token RAG prompt with the following structure (sentences abbreviated):

\begin{quote}
\small
\textbf{[1]} You are a helpful financial assistant. \textbf{[2]}~Answer questions using only the provided context. \textbf{[3]}~If the answer is not in the context, say ``I don't know.'' \textbf{[4]}~Context: The company reported Q3 revenue of \$4.2B\ldots\ \textbf{[5]}~Operating expenses increased 12\% year-over-year\ldots\ \textbf{[6]}~The board approved a \$500M share buyback program\ldots\ \textbf{[7]}~Employee headcount remained stable at 12{,}400\ldots\ \textbf{[8]}~Regional sales in APAC grew 8\%\ldots\ \textbf{[9]}~The CFO noted that ``margins are expected to improve in Q4.''\ldots\ \textbf{[10]}~\ldots\emph{(180 more sentences of retrieved context)}\ldots\ \textbf{[190]}~Question: What is the company's revenue outlook?
\end{quote}

\paragraph{Step 1: Sentence Segmentation.}
The prompt is split into 190 sentences using the multilingual rule-based segmenter.

\paragraph{Step 2: Scoring.}
Each sentence receives a composite score combining TextRank, position, TF-IDF, and novelty signals (\S\ref{sec:compression}).

\begin{itemize}
\item \textbf{Position weighting} assigns high scores to sentences [1]--[4] (system prompt, instructions) and [190] (the user question) via the U-shaped curve; middle sentences [90]--[100] receive the lowest weight.
\item \textbf{TextRank} identifies [4], [6], and [9] as central (they share financial terms with many other sentences).
\item \textbf{TF-IDF} upweights [6] (``buyback'' is rare/distinctive) and [9] (direct CFO quote with ``margins'', ``outlook'').
\item \textbf{Novelty} would upweight any sentence whose vocabulary diverges from the financial-report average---e.g., an injected jailbreak prefix or PII amid earnings data would score high.
\end{itemize}

\paragraph{Step 3: Selection and Reordering.}
Sentences [1]--[3] and [189]--[190] are always preserved (primacy/recency). The remaining budget (${\sim}$512 tokens minus preserved sentences) is filled by top-ranked sentences. The selected sentences are reordered by original position:

\begin{quote}
\small
\textbf{[1]} You are a helpful financial assistant. \textbf{[2]}~Answer questions using only the provided context. \textbf{[3]}~If the answer is not in the context, say ``I don't know.'' \textbf{[4]}~Context: The company reported Q3 revenue of \$4.2B\ldots\ \textbf{[6]}~The board approved a \$500M share buyback program\ldots\ \textbf{[9]}~The CFO noted that ``margins are expected to improve in Q4.''\ldots\ \textbf{[190]}~Question: What is the company's revenue outlook?
\end{quote}

The compressed prompt retains the system instructions, the most informative context sentences, and the user question in their original order---reducing ${\sim}$2{,}048 tokens to ${\sim}$480 tokens (4.3$\times$ compression) while preserving all content relevant to domain classification, jailbreak detection, and PII scanning. The original, uncompressed prompt is forwarded to the upstream LLM; only the router's classifiers see the compressed version.

\end{document}